# American Sign Language Identification Using Hand Trackpoint Analysis

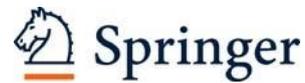

Yugam Bajaj[1], Puru Malhotra[*1]

[1] Dept. of Information Technology, Maharaja Agrasen Institute of Technology, Delhi, India
{ yugambajaj99, purumalhotra99}@gmail.com

**Abstract**

Sign Language helps people with Speaking and Hearing Disabilities communicate with others efficiently. Sign Language identification is a challenging area in the field of computer vision and recent developments have been able to achieve near perfect results for the task, though some challenges are yet to be solved. In this paper we propose a novel machine learning based pipeline for American Sign Language identification using hand track points. We convert a hand gesture into a series of hand track point coordinates that serve as an input to our system. In order to make the solution more efficient, we experimented with 28 different combinations of pre-processing techniques, each run on three different machine learning algorithms namely k-Nearest Neighbours, Random Forests and a Neural Network. Their performance was contrasted to determine the best pre-processing scheme and algorithm pair. Our system achieved an Accuracy of 95.66% to identify American sign language gestures.

## 1. Introduction

American Sign Language (ASL) uses hand gestures and movements as a means of communication for people with hearing or speaking disabilities. It is a globally recognized standard for sign language but still, there are only ~250,000 – 500,000 people who understand it [1] and this makes the users dependent on ASL restricted while conversing in real-life scenarios. In this paper, we propose an ASL recognition system to tackle this problem and hence lay a foundation for translator devices to make dynamic conversation in ASL easier. The proposed system uses a series of pre-processing steps to convert a gesture image into meaningful numeric data using hand TrackPoint analysis which serves as the input to a Machine Learning/Deep Learning algorithm for identification. The existing solutions require the use of external devices such as motion sensing gloves or Microsoft Kinect to capture the essence of finger movements. Hence, decreasing the feasibility and acces-sibility. Also, they use a really complex system of Neural Networks which is computationally expensive.

We propose an American Sign Language gesture identification method which is capable of recognising 24 Alphabet letters. Our system is based on the pipeline as shown in Figure 1. A Gesture image goes through it while being converted to a series of hand track points. So, our system represents a particular ASL gesture as coordinates (track points) instead of an image. In the study, various pre-processing combinations were compared by their performance on k-Nearest Neighbour Algorithm, Random Forest Algorithm and a 9 Layered Sequential Neural Network to find the best possible pre-processing + algorithm pair. Our study was able to achieve the state-of-the-art results, with less complicated algorithms from the ones that are commonly used.

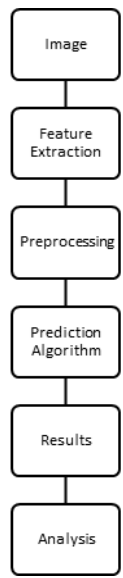

**Fig. 1:** Process Pipeline

## 2.  Related Work

A lot of research has been done till now to solve the problem of sign language detection, the majority of them applying Neural Networks and were able to achieve accuracies greater than 90% [2-11].

In one such experiment, Razieh Rastgoo et al. constructed multi-view skeletons of the hands from video samples and fed them into a 3D-Convolutional Neural Network (CNN) model. They outperformed state of the art models on New York University (NYU) and First-Person datasets. They applied 3DCNN on stacked input to get discriminant local Spatio-temporal fea-tures and the outputs were fused and fed to a Long Short-Term Memory (LSTM) Network which formulated different hand gestures [12].

Zafar Ahmed Ansari and Gaurav Harit made an effort to solve the problem of identification of Indian Sign Language. Indian Sign Language, being more complex to tackle as a Computer vision task than other Sign Language standards due to the use of both hands and involving movement, has been less explored as compared to advancements made in recognition methods for other sign languages. In their paper titled 'Nearest neighbour classification of Indian sign language gestures using Kinect camera', they were able to achieve an accuracy of 90.68% for the task. They used a k Nearest neighbour (kNN) approach as it provides a good bases for such a task where samples can be from different angles and backgrounds [13].

## 3. Data Collection

The data samples were collected through a dummy web interface developed using Java Script and HTML which was hosted on a local web server. The website platform supported functionality to capture images, process them and record the hand trackpoints as results after feature extraction. The images were captured using a standard Webcam integrated with a laptop. The samples were collected from 4 volunteers who performed the hand gestures to be captured by the webcam with their hands at a distance ranging between 1.5 feet to 3 feet from the webcam. This distance was sufficient enough to capture the full gesture and was a good estimate to mimic the distance between two people in a conversation in real life. The hand gestures corresponding to 24 of American Sign Language Alphabets were performed, omitting gestures for the letters J and Z as they involved movement. 30 samples for each alphabet were captured per volunteer which compiled to a total of 120 samples per alphabet. The backgrounds of the image while capturing the samples were kept a mix of plain background with only the hand in the frame and a natural background with the person performing the gesture in the frame.

Figure 2 shows samples from our collected data corresponding to each of the alphabets mentioned. Each of such images were processed to be converted into 21 3-dimensional coordinate points as per the process mentioned in the next sub section and then stored in an ordered manner in a csv file to be used as a dataset to a Machine Learning model. The dataset had 64 columns – first 63 containing the $n_x$ column followed by $n_y$ column followed by the $n_z$ column representing the x, y and z coordinates for trackpoint $n$, where $n \epsilon [1, 21]$ in order and column 64 containing the actual representation of the gesture in English.

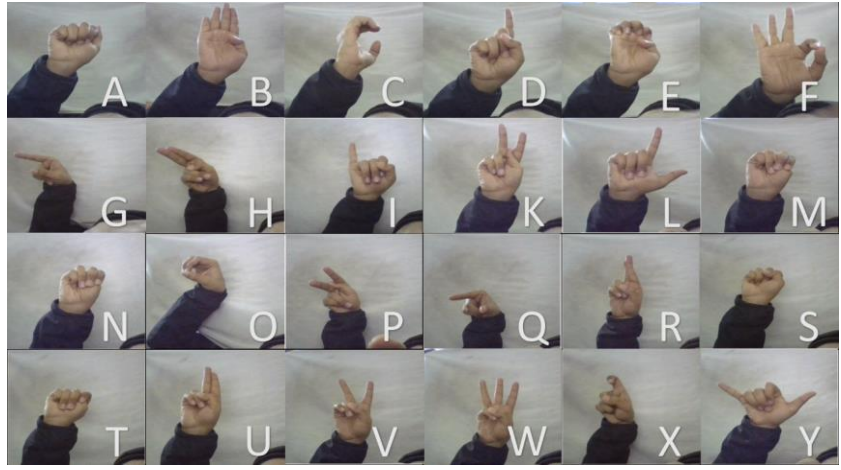

**Fig. 2**: Dataset Sample: American sign language chart

### 3.1. Feature Extraction

The captured images were passed to a Hand tracking model from the MediaPipe framework which was deployed using tensorflow js. The Hand tracking model's parameters and weights were adjusted as per our experiment's requirements. The Hand Tracking model uses a BlazePalm detector to detect the palm in the captured image and passes the section of the image with the palm in it to a Hand Landmark Model [14]. The segment for the feature extraction process returned a Tensor of float values which contained the coordinates of 21 Landmark points of a Hand in 3D space. These coordinates serve as the data values for further carrying out the task of American Sign Language recognition.

## 4. Preprocessing

### 4.1. Normalization of Coordinates

The Hand Gesture can be present anywhere in the Image frame. The position of the hand in the image affects the reference point coordinates for all the 3 (x,y,z) axes. To make it easy for the algorithms to generalize patterns for different gestures, we have to bring all of them to a single reference frame. For that, all the coordinates of a particular image sample were imagined to be contained within a Bounding Box, which was taken as a reference for calculations.

In our study, we carried out all our calculations with two types of Bounding Boxes:

**Cuboidal Bounding Box**. This type of Bounding Box was the smallest cuboid that could fit in all the 21 reference points in the space. For each image the following 6 points were found: $x_{min}$, $x_{max}$, $y_{min}$, $y_{max}$, $z_{min}$, $z_{max}$; which denoted the minimum and maximum coordinate value for their respective axis.

Hence the bounding box stretched from $x_{min}$ to $x_{max}$, $y_{min}$ to $y_{max}$ and $z_{min}$ to $z_{max}$ in the x, y, z axis respectively and was of the dimensions $(x_{max} - x_{min}) * (y_{max} - y_{min}) * (z_{max} - z_{min})$.

**Cubical Bounding Box**. This type of Bounding box was the smallest cubical box that could fit in all the 21 reference points in the space. For each image all the 6 points for the minimum and maximum points for each axis was found as above. Now the largest value amongst $(x_{max} - x_{min})$, $(y_{max} - y_{min})$ and $(z_{max} - z_{min})$ acted as the length of the edge for the cube and the other two lengths were made equal to it by uniformly adjusting their min and max value. This was done using the following general equations:

$$T_{min(new)} = T_{min(old)} - \frac{Edge\ length\ of\ cube - (T_{max} - T_{min})}{2}. \quad (1)$$

$$T_{max(new)} = T_{max(old)} + \frac{Edge\ length\ of\ cube - (T_{max} - T_{min})}{2}. \quad (2)$$

For any axis T.

Hence the bounding box stretched from modified $x_{min}$ to $x_{max}$, $y_{min}$ to $y_{max}$ and $z_{min}$ to $z_{max}$ in the x, y, z axis respectively and was of the edge length $(x_{max} - x_{min})$.

For normalizing the coordinates into a common reference frame, we use two mathematical transformations on our data:

- Shifting of Origin
- Scaling

**Shifting of Origin**. The reference vertex of the Bounding Box was shifted back to the origin to bring all the coordinates to a common reference position. The reference vertex of the Bounding Box was considered to be the vertex $(x_{min}, y_{min}, z_{min})$ for the box. The shifting of each coordinate took place following the simple mathematical transformation for shifting of origin:

$$X = x - x_{min}. \tag{3}$$

$$Y = y - y_{min}. \tag{4}$$

$$Z = z - z_{min}. \tag{5}$$

For every coordinate point (x,y,z)

**Scaling.** Each Bounding box is scaled to a standard size of (255 * 255 * 255). This is done to bring every collected image sample to a uniform size for the box, bringing every coordinate placed in a common reference space. This was aimed to increase the efficiency of classification algorithms because of the ease of comparison of the generalization for patterns for every gesture. A Scaling Factor(f) was calculated for each dimension of the bounding box by using the formula:

$$f = \frac{255}{L}. \tag{6}$$

where L indicates the edge length of the bounding box in each of the x, y and z direction one by one.

Each coordinate value was multiplied by the scaling factor for their respective axis in order to complete their transformation into a box of the required dimensions.

### 4.2. Rounding Off

Another technique which was used to pre-process the coordinate values was the process of rounding up the values. After many trials and experimenting, it was found that rounding up the values to 3 decimal places produced consistent results. Hence rounding coordinates values to 3 places after the decimal was performed in the study wherever applicable.

## 5. Method

Three different algorithms were used in the experiment to build a sign language recognition model and their performance were tested on 28 different combinations of the discussed pre-processing techniques. These algorithms were:

- k-Nearest Neighbour Classifier

- Random Forest Classifier
- Neural Network

The collected Dataset was distributed randomly into an 80:20 split of Training: Test set for the training and testing purpose of the models.

### 5.1. k Nearest Neighbours Classifier (kNN)

For a data record $t$ to be classified using kNN, its $k$ nearest neighbours are retrieved, and this forms a neighbourhood of $t$ [15]. Hence the record $t$ is labelled to be belonging to the category of the corresponding neighbourhood. The performance of the algorithm is largely dependent on the choice of $k$. Hence, to find the optimal value of $k$, the classifier was tested for different values of $k$ in the range (1,25).

### 5.2. Random Forest Classifier

Random Forest Classifier uses a series of decision trees to label a sample. The performance of the algorithm depends upon the number of decision trees($n$) which the algorithm forms to make a prediction. To find the optimal value of the number of decision trees the classifier was tested for different values of $n$ in the range (1,200).

### 5.3. Neural Network

A Neural Network was constructed in keras to carry out the experiment. The network contained 9 Dense layers with the initial 8 layers powered by the Relu activation function and the final output layer powered by the Softmax activation function which outputs the probabilities of the input belonging to each of the available labels. The network was compiled using a Categorical Cross Entropy Loss function with an Adam Optimizer and an Accuracy metrics. For its training purpose, the neural network was fitted to the training set and was trained for 128 epochs.

## 6. Observations

The recognition system was implemented on an Intel Core i5 CPU (2.40 GHz × 4) and NVIDIA Geforce GTX 1650 GPU with 8 GB RAM. The system ran Windows 10 (64 bit). The system was implemented in Python programming language. Matplotlib and Seaborn package was used for analysis and visualising results. Table 1 shows the results obtained by using a cuboidal bounding box and a cubical bounding box.

Table 1. Test Set Accuracy using a cuboidal and cubical bounding box over all the pre-processing combinations.

| Pre-processing Combination | Cuboidal Bounding Box | | | Cubical Bounding Box | | |
|---|---|---|---|---|---|---|
| | kNN | Random Forest | Neural Network | kNN | Random Forest | Neural Network |
| No Pre-processing | 70.83 | 77.43 | 91.49 | 70.83 | 77.43 | 87.33 |
| Shifting | 90.97 | 91.15 | 93.06 | 90.97 | 91.15 | 94.27 |
| Scaling | 67.53 | 76.91 | 89.06 | 69.1 | 75.35 | 81.94 |
| Rounding | 70.83 | 77.08 | 87.5 | 70.83 | 70.83 | 91.15 |
| Scaling + Shifting | 92.71 | 93.23 | 92.88 | **93.23** | 92.88 | 93.92 |
| Scaling + Rounding | 67.53 | 78.3 | 87.85 | 69.1 | 75.35 | 89.58 |
| Rounding + Scaling | 67.53 | 77.6 | 84.9 | 69.1 | 74.13 | 88.54 |
| Shifting + Rounding | 90.97 | 91.32 | 94.97 | 90.97 | 91.49 | 93.92 |
| Rounding + Shifting | 90.97 | 91.32 | 93.23 | 90.97 | 91.32 | 94.44 |
| Rounding + Shifting + Scaling | 92.71 | **93.4** | 93.4 | **93.23** | 93.23 | **95.66** |
| Shifting + Scaling + Rounding | 92.71 | 93.23 | 88.72 | **93.23** | 92.88 | 91.67 |
| Rounding + Scaling + Rounding | 67.53 | 77.26 | 90.45 | 69.1 | 75.17 | 83.51 |
| Rounding + Shifting + Rounding | 90.97 | 91.32 | 93.92 | 90.97 | 91.32 | 94.62 |
| Rounding + Shifting + Scaling + Rounding | 92.71 | **93.4** | 91.32 | **93.23** | 92.88 | 93.4 |

Both the types of bounding boxes performed relatively well over the set of different combinations of pre-processing techniques. Among the pre-processing combinations, the combination of Rounding the data first and then performing Shifting and Scaling respectively resulted in giving maximum accuracy. The performance results for each algorithm are further discussed.

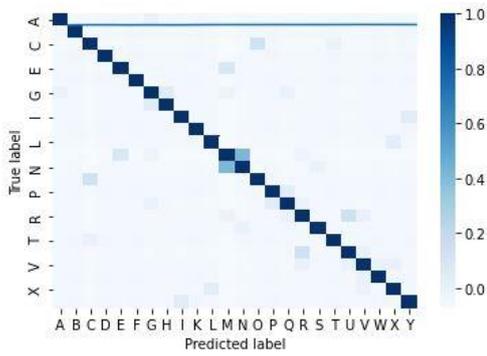

**Fig. 3**: Confusion matrix for best performance in kNN

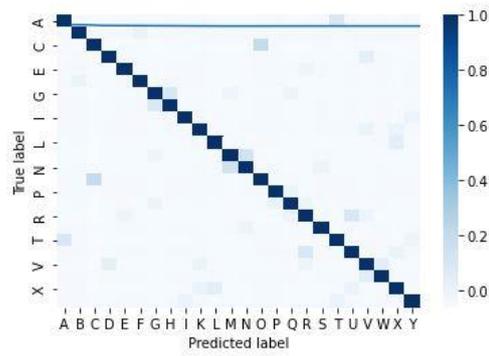

**Fig. 4**: Confusion matrix for best performance in Random Forest

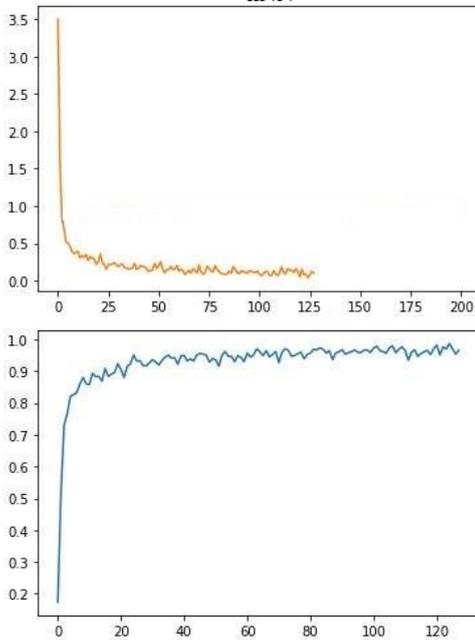

**Fig. 5**: Loss and Training accuracy curve for Neural network

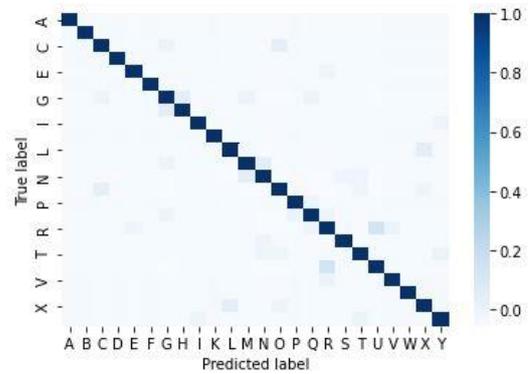

**Fig. 6**: Confusion matrix for best performance in Neural Network

### 6.1. kNN

The kNN Algorithm performed with an average accuracy of 82.19% over the 28 experimental combinations. Its best performance recorded was for four different combinations which had an accuracy of 93.23%. These combinations were namely:

- Shifting + Scaling
- Rounding + Shifting + Scaling
- Shifting + Scaling + Rounding
- Rounding + Shifting + Scaling + Rounding

all in a cubical bounding box. All the methods achieving maximum accuracy had similar results, recognizing 8 Characters with 100% accuracy and 10 other characters with accuracies over 90% (as shown in figure 3).

### 6.2. Random Forest

The Random Forest Algorithm gave an average accuracy of 85.30% over the 28 experimental combinations. Its best accuracy of 93.4% was achieved for the pre-processing combination of:

- Rounding + Shifting + Rounding
- Rounding + Shifting + Scaling + Rounding

Both over a Cuboidal Bounding Box. Both of them recognized 7 letters with 100% accuracy and 12 others with accuracies over 90% (as shown in figure 4).

### 6.3. Neural Network

The Neural Network architecture designed by us performed the best with an average accuracy of 90.95%. A combination involving a pre-processing combination of Rounding + Scaling + Shifting over a Cubical Bounding Box recorded an accuracy of 95.66%, which was the maximum accuracy achieved over the experiment, compiled with a loss < 0.1 and Training Accuracy of 0.978 (figure 5). The Neural Network recognized 11 characters with 100% accuracy taking the total to 20 out of 24 letters recognised with accuracies above 90% (as shown in figure 6).

## 7. Results and Discussion

The study was focused on developing a system to detect American Sign Language gestures. In the course of the study, there was a comparison made between the kNN Algorithm, Random Forest Classifier and a proposed Neural Network. The kNN algorithm performed with an accuracy of 70.83%, Random forest performed with an accuracy of 77.43% and the Neural Network with an accuracy of 91.49% on raw unprocessed data. To find an optimised solution we tried 28 different pre-processing techniques. The use of pre-processing helped increase the performance of the 3 algorithms to a maximum of 93.23%, 93.4% and 95.66% respectively. A comparative analysis of the study indicates the Neural Network being the most effective with an average accuracy of 90.95%, outperforming kNN and Random Forest which had the average accuracies of 82.19% and 85.30% respectively over a total of 28 test runs completed during the study. We concluded by achieving a maximum Accuracy of 95.66% over our Test Dataset in a combination using a Neural Network. Among all the pre-processing techniques, the combination applying Rounding + Shifting + Scaling proved out to be the most efficient giving out high accuracies in all the 3 algorithms. Hence, a pipeline was devised to serve as an American Sign Language identification system.

## 8. Conclusion and Future Work

We Implemented and Trained an American Sign Language identification system. We were able to produce a robust model for letters a, b, d, e, f, i, k, l, o, s, x and a modest one for letters a-y (except r). The pre-processing techniques implemented resulted in a 14.47% average increase on test set accuracy over no pre-processing.

This work can be further extended to develop a two-way communication system between Sign Language and English. Also, there is a need to further work out finding ways to recognize letters or Gestures involving hand moments as well.